\def\year{2019}\relax
\documentclass[letterpaper]{article} 
\usepackage{aaai19}  
\usepackage{times}  
\usepackage{helvet}  
\usepackage{courier}  
\usepackage{url}  
\usepackage{graphicx}  
\usepackage{amsmath}
\begin{document}
%
\title{Video Inpainting by Jointly Learning Temporal Structure and Spatial Details}

\author{Chuan Wang$^1$ $\quad$ Haibin Huang$^1$ $\quad$ Xiaoguang Han$^{\star2}$ $\quad$Jue Wang$^1$ \\ \\
Megvii (Face++)$^1$ \\
\texttt{\{wangchuan, huanghaibin, wangjue\}@megvii.com} \\
Shenzhen Research Inst. of Big Data, The Chinese University of Hong Kong, Shenzhen, China$^2$ \\
\texttt{hanxiaoguang@cuhk.edu.cn} \\
}

\maketitle
\let\thefootnote\relax\footnotetext{$^\star$Corresponding Author.}

\begin{abstract}
We present a new data-driven video inpainting method for recovering missing regions of video frames. A novel deep learning architecture is proposed which contains two sub-networks: a temporal structure inference network and a spatial detail recovering network. The temporal structure inference network is built upon a 3D fully convolutional architecture: it only learns to complete a low-resolution video volume given the expensive computational cost of 3D convolution. The low resolution result provides temporal guidance to the spatial detail recovering network, which performs image-based inpainting with a 2D fully convolutional network to produce recovered video frames in their original resolution. Such two-step network design ensures both the spatial quality of each frame and the temporal coherence across frames. Our method jointly trains both sub-networks in an end-to-end manner. We provide qualitative and quantitative evaluation on three datasets, demonstrating that our method outperforms previous learning-based video inpainting methods.
\end{abstract}

\section{Introduction}\label{sec:intro}
\begin{figure} [t]
  \centering
  \includegraphics[width=0.95\linewidth]{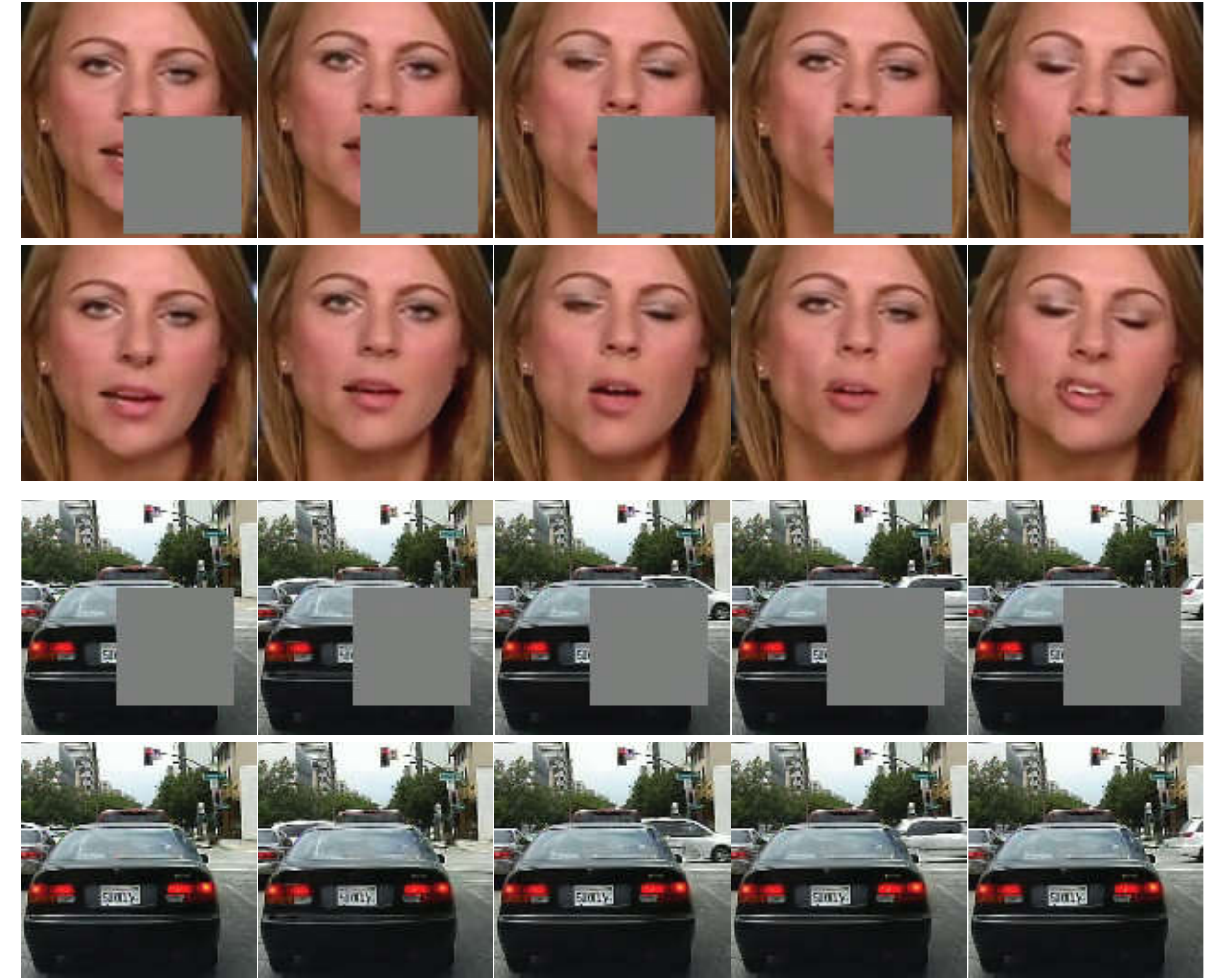}
  \caption{Video inpainting results by our approach. Row 1, 3: input frames with missing regions (shown in gray). Row 2, 4: our results. Note that the filled regions contain rich image details and are temporally coherent.}\label{fig:teaser} 
\end{figure}

Given an image or a video with holes inside (e.g. generated by object removal), inpainting (also called completion) techniques try to recover the missing video content to produce a natural looking result. This problem has drawn great attention in the past two decades due to strong industrial demands on image and video editing applications. Inpainting is a very challenging task as the requirements are two-fold: (1) the generated content in the missing regions must be semantically correct given their surrounding  content; and (2) the missing regions need to be filled in a seamless way so that the original holes are visually unnoticeable.

In this work, we focus on video inpainting, an extended problem from image inpainting with the added temporal dimension. Such extension brings new technical challenges that are difficult to resolve. First, recovering  missing video content  requires the understanding of not only the spatial context of each frame, but also the motion context across frames. Second, the output video must maintain high spatio-temporal consistency, in both global context-level and local image-feature-level. Although there has been tremendous progress in image inpainting, e.g. patch-based image synthesis~\cite{barnes2009patchmatch,barnes2010generalized} and deep learning based approaches~\cite{pathak2016context,IizukaSIGGRAPH2017}, a direct extension of those methods in 3D would not work well for video inpainting. Specifically, \cite{wexler2007space,venkatesh2009efficient,newson2014video,huang2016temporally} have tried to extend spatial 2D patch synthesis to spatial-temporal 3D patch synthesis. However, local synthesis, even in 3D space, cannot guarantee global semantic correctness. 
Moreover, directly applying image inpainting networks to each frame individually often leads to temporally jittering results that are visually unacceptable, as we will demonstrate later.

\begin{figure*}[t!]
  \centering
  \includegraphics[width=0.95\linewidth]{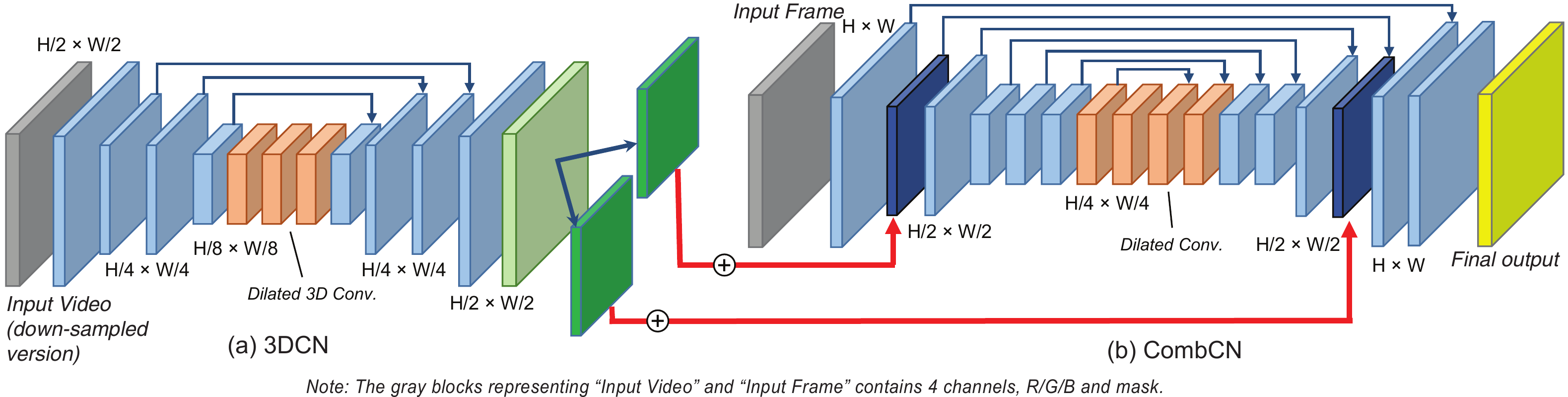} 
  \caption{Network architecture of our 3D completion network (3DCN) and 3D-2D combined completion network (CombCN). The 3DCN works in low resolution, producing an inpainted video as output. Its individual frames are further convolved and added into the first and last layer of the same size in CombCN. The input video for 3DCN and the input frame for CombCN, shown as gray blocks, are in 4-channel format, containing RGB and the mask indicating the holes to be filled.}\label{fig:net3d}
\end{figure*}

We present a novel end-to-end deep learning architecture to tackle the above issues for high quality video inpainting.
As shown in Fig.~\ref{fig:net3d}, the network consists of a temporal structure prediction sub-network and a spatial detail recovering sub-network. The former sub-network treats a video as a 3D volume.  It takes a down-sampled version of the original video as input, and fill the holes in it using 3D CNN with an Encoder-Decoder architecture. We use this output volume as temporal structure guidance, since it captures the motion structure across time but lacks spatial details. The spatial detail inference network then takes the original video and the temporal structure guidance as input, and generates completed video frames in their original resolution. It has a 2D Encoder-Decoder architecture with global and local $l_1$ consistency losses. These two sub-networks are jointly trained and can benefit from each other. In other words, the temporal structure guidance improves both the temporal smoothness and the context consistency of the final video. Meanwhile, the loss of the spatial detail recovering network is also back-propagated into the first network and helps improve the accuracy of temporal structure prediction.


In summary, our main contributions are:
\begin{itemize}
\setlength\itemsep{-0.1em}
    \item it is the first work to use deep neural networks for solving the problem of video completion. Compared with existing methods, the proposed algorithm can deal with the video with complex appearances and large missing regions;
    \item we design a novel deep learning architecture that uses 3D CNN for temporal structure prediction and 2D CNN for spatial detail recovering, where the output temporal structure is fused into the 2D CNN to guide the detail inference;
    \item we perform joint training of the two sub-networks, which further improves the performance of the overall system.
\end{itemize}


\section{Related Work}\label{sec:related}
In this section, we introduce the related works in the following three aspects and refer the readers to the works of~\cite{chhabra2014detailed} and~\cite{ilan2015survey} for detailed literature review of image/video inpainting. 

\paragraph{Patch-based image/video inpainting.} To fill in the holes using patch-based synthesis is the most used traditional strategy for image inpainting. This
was firstly proposed in~\cite{efros1999texture}, where the missing contents are recovered in a region-growing way: the method starts
from boundary of holes and extends the region by searching appropriate patches and assembling them together. Following up this work, there are many different directions for improvement in searching and optimization~\cite{kwatra2005texture,wexler2007space,barnes2009patchmatch,barnes2015patchtable}, or for application like face~\cite{zhao2018identity,yamaguchi2018high}. 
It is also adapted to video inpainting problem by replacing 2D patch synthesis with 3D spatial-temporal patch synthesis across frames. This
was firstly proposed in~\cite{wexler2004space,wexler2007space} to ensure the temporal consistency of the generated video and later improved in~\cite{jia2005video,venkatesh2009efficient} to handle more complicated video input.  However, all of these works are designed for the video with repeated content across frames. They are unable to tackle the problem we proposed in this paper where missing parts cannot be replaced by similar content in the input. Resorting to a large video dataset, we try to train a CNN in this work for missing contents prediction based on the high-level context
understanding. 

\paragraph{Image completion using 2D CNN.} Recently, Convolutional Neural Network was firstly used in~\cite{xie2012image} for image inpainting but only for small holes.
Pathak et al.~\cite{pathak2016context} then proposed to deal with large missing regions using an encoder-decoder architecture which can efficiently learn the
context feature of the image. For high-resolution image inpainting, Yang et al.~\cite{yang2017high} developed a multi-scale neural patch synthesis algorithm that not only preserves
contextual structure but also produces high-frequency details. The algorithm proposed in~\cite{IizukaSIGGRAPH2017} further improves the performance by involving
two adversarial losses to measure both the global and local consistency of the result. Different from the previous works which only focus on box-shaped holes, it
also develops a strategy to handle the holes with arbitrary shapes. To extend these methods from image to video domain is a challenging task, because video
completion not only needs to have an accurate context understanding of both frames and motions, but also requires to ensure temporal smoothness of the
output. In this paper, we propose a novel deep learning architecture for this problem which takes use of both 2D and 3D CNNs to jointly learn
the temporal structure and spatial details. 
\paragraph{Shape completion using 3D CNN.} Another series of works related to our paper is using 3D CNN for 3D shape completion. Similar to deep learning
based image inpainting, most of the methods such as~\cite{sharma2016vconv,varley2017shape,dai2017shape} uses encoder-decoder architecture but with
3D CNN for solving this problem. However, all these techniques can only handle low-resolution grids (typically $30^3$ voxels) due to the high computational
cost of 3D convolutions. To address this issue, many approaches are proposed most recently. Resorting to a dataset, Dai et al.~\cite{dai2017shape} used patch
retrieval and assembly as a post-processing to refine the low-resolution output of encoder-decoder network. For such post-refinement, the method in~\cite{wang2017shape} proposed a strategy to slice the low-resolution output into a sequence of images and did super-resolution and completion for each sliced image with a recurrent neural network. Han et al.~\cite{HRSC} designed a hybrid networks for jointly global structure prediction and local geometry inference. Our work is inspired by this method but differs from it in two aspects. Firstly, the method of~\cite{HRSC} conducted the completion in a
region-growing way while ours is an end-to-end architecture for completion. Secondly, for details inference, the method in~\cite{HRSC} only looks at a local
region which lacks much surrounding context information while our algorithm uses the content of the whole image for missing information recovering.

\section{Algorithm}\label{sec:na}
Our method is built upon deep neural networks. It takes an incomplete video $V_{in}$ and a mask video $M$ as input, and produces an complete video $V_{out}$ as output. The incomplete video is represented as a $F\times H\times W$ volume, where $F$, $H$, $W$ are number of frames, height and width of $V_{in}$. $M$ and $V_{out}$ are of the same size as $V_{in}$. In training stage, $V_{in}$ is obtained by randomly generating holes on a complete video $V_{c}$, i.e. the target ground truth of the inpainting task.

To produce a spatial-temporal coherent inpainted video, our network consists of two sub-networks. One network is a 3D completion network (3DCN), which utilizes 3D CNN to infer the temporal structure globally from a down-sampled version ($F\times \frac{H}{r}\times \frac{W}{r}$) of $V_{in}$ and $M$. In this paper, we use superscript $d$ to distinguish the downsampled videos from their original versions. So 3DCN takes $V^d_{in}$ and $M^d$ as input and produces $V^d_{out}$ as output. Another sub-network is a 3D-2D combined completion network (CombCN). It applies 2D CNN to perform completion frame by frame. The input of this network includes one incomplete but high-resolution frame of $V_{in}$ with its mask frame of $M$, and a low-resolution but complete frame $I^{d,k}_{out}$ of $V^d_{out}$ ($k = 1,2,...,F$). In our paper, $F$, $H$, $W$, $r$ are set to 32, 128, 128 and 2 respectively.

\begin{table*}[t]
\centering
\resizebox{0.85\linewidth}{!}{
\begin{tabular}{c|ccccc||c|ccccc}
\hline
Layer No. & Type          & Kernel & Stride & Channel & Dilation & Layer No. & Type          & Kernel & Stride & Channel & Dilation \\ \hline
1         & conv.         & 5      & 1      & 16      & -        & 7         & dilated conv. & 3      & 1      & 256     & 8        \\
2         & conv. $\downarrow$         & 3      & 2      & 32      & -        & 8         & conv.         & 3      & 1      & 128     & -        \\
3         & conv.         & 3      & 1      & 64      & -        & 9         & deconv. $\uparrow$       & 4      & 2      & 64      & -        \\
4         & conv. $\downarrow$         & 3      & 2      & 128     & -        & 10        & conv.         & 3      & 1      & 32      & -        \\
5         & dilated conv. & 3      & 1      & 256     & 2        & 11        & deconv. $\uparrow$      & 4      & 2      & 16      & -        \\
6         & dilated conv. & 3      & 1      & 256     & 4        & 12        & conv.         & 3      & 1      & 3       & -        \\ \hline
\end{tabular}}\\
\resizebox{0.85\linewidth}{!}{
\begin{tabular}{c|ccccc||c|ccccc}
\hline
Layer No. & Type          & Kernel & Stride & Channel & Dilation & Layer No. & Type          & Kernel & Stride & Channel & Dilation \\ \hline
1         & conv.         & 5      & 1      & 64      & -        & 10        & dilated conv. & 3      & 1      & 256     & 16       \\
2$^*$         & conv. $\downarrow$        & 3      & 2      & 128     & -        & 11        & conv.         & 3      & 1      & 256     & -        \\
3         & conv.         & 3      & 1      & 128     & -        & 12        & conv.         & 3      & 1      & 256     & -        \\
4         & conv. $\downarrow$         & 3      & 2      & 256     & -        & 13        & deconv. $\uparrow$       & 4      & 2      & 128     & -        \\
5         & conv.         & 3      & 1      & 256     & -        & 14        & conv.         & 3      & 1      & 128     & -        \\
6         & conv.         & 3      & 1      & 256     & -        & 15$^*$        & deconv. $\uparrow$       & 4      & 2      & 64      & -        \\
7         & dilated conv. & 3      & 1      & 256     & 2        & 16        & conv.         & 3      & 1      & 32      & -        \\
8         & dilated conv. & 3      & 1      & 256     & 4        & 17        & conv.         & 3      & 1      & 3       & -        \\
9         & dilated conv. & 3      & 1      & 256     & 8        &         &               &        &        &         &          \\ \hline
\end{tabular}}\vspace{2mm}
\caption{Network architecture of 3DCN (top) and CombCN (bottom). In the bottom table, * represents the layers where combination takes place in CombCN.} \label{tab:network-arch-3D-and-combined}
\vspace{-4mm}
\end{table*}

\subsection{Temporal structure inference by 3DCN}
A video inpainting task requires not only filling the missing part within each frame but also keeping the consistency between successive frames. For this reason, it would fail if directly apply the existing image inpainting methods such as~\cite{IizukaSIGGRAPH2017} to video frames separately, because they lack the mechanism of preserving temporal coherence. On the other hand, video can also be viewed as a spatial-temporal volume and its temporal structure could be preserved when algorithms are globally applied to it, such as~\cite{wang2017video,hara3dcnns,wang2014video}. To this end, we apply 3D CNN globally to video inpainting. However, due to the expensive memory cost of 3D convolutions, we only utilize 3D CNN on a down-sampled version $V^d_{in}$ of the input video. The 3D completion network can generate an inpainted video $V^d_{out}$ which captures the temporal structure of the original video, even though its individual frame lacks details.



Our 3D completion network follows an encoder-decoder structure and consists of 12 layers totally. Given the incomplete video and its mask as input, it first exploits 4 strided convolutional layers to encode it to a latent space, capturing its temporal-spatial structure. Then 3 dilated convolutional layers with rate 2, 4, 8 are followed to capture the spatial-temporal information in larger perception field. At last, video inpainting is finally achieved by 3 convolutional and 2 fractionally-strided convolutional layers in an alternative order, yielding the result with missing part filled. Rather than using the max-pooling and upsampling layers to compute the feature maps, we employ $3\times3$ convolution kernels with stride of 2, which ensures that every pixel contributes. Meanwhile, considering that non-successive frames may have loose relations with the current one, and to avoid information loss across frames, we limit stride and dilation to take effects only within frame rather than across frames. As a result, the feature map of each layer has a constant frame number $F$. Besides, all the convolutional layers are followed by batch normalization (BN) and ReLU non-linearity activation except the last one. Paddings are involved to make the input and output have exactly the same size. The skip-connections as U-Net~\cite{ronneberger2015u} are also employed to facilitate the feature mixture across encoder and decoder. 
The detailed configuration of our 3D completion network is illustrated in Fig.~\ref{fig:net3d} (a) and listed in Table~\ref{tab:network-arch-3D-and-combined} (top) (BN and ReLU are not shown for brevity). 

\paragraph{Training.} Let $G_v(V_{in}^d, M^d) = V^d_{out}$ denotes the 3DCN in a functional form. The binary masks $M^d$ and $M$ take the value 1 inside regions to be filled-in and 0 elsewhere. The pixels of $V^d_{in}$ and $V_{in}$ inside the mask region are pre-filled with the mean pixel value of the training dataset before feeding it to the network. During training, we minimize the $l_1$ norm the difference between $V^d_{out}$ and $V_c^d$. The difference is also weighted considering the completion region mask is used. Specifically, the $l_1$ loss of 3DCN is defined by: 
\begin{equation}\label{eq:l1-3dcn}
L^{3DCN}(V_{in}^d, M^d, V_c^d) = \frac{\lVert M^d \odot (G_v(V_{in}^d, M^d) - V_c^d) \rVert}{\lVert M^d \rVert}
\end{equation}
where $\odot$ is the pixelwise multiplication and $\lVert\cdot\rVert$ is the $l_1$ norm.

\subsection{Spatial details inference by CombCN}
The output of 3DCN is a low resolution inpainted video. It conveys temporal structure but lacks details within each frame. To restore the details, we extend a 2D completion network (2DCN) inspired by a state-of-the-art image inpainting work~\cite{IizukaSIGGRAPH2017}, obtaining a combined completion network (CombCN). This CombCN consists of 17 layers including 11 strided convolutional layers, 2 fractional deconvolutional layers and 4 dilated convolutional layers. It also follows an encoder-decoder structure where the minimal feature map size is $\frac{H}{4}\times\frac{W}{4}$. The dilated convolutional layers are involved to obtain a larger perception field so that the network can "see" areas far from the missing part. The configuration of CombCN is listed in Table.~\ref{tab:network-arch-3D-and-combined} (bottom) and we encourage the readers to review~\cite{IizukaSIGGRAPH2017} for a detailed explanation of its original configuration. Note that we also made a modification by involving skip-connection as U-Net.

\begin{figure*}[t!]
  \centering
  \includegraphics[width=0.92\linewidth]{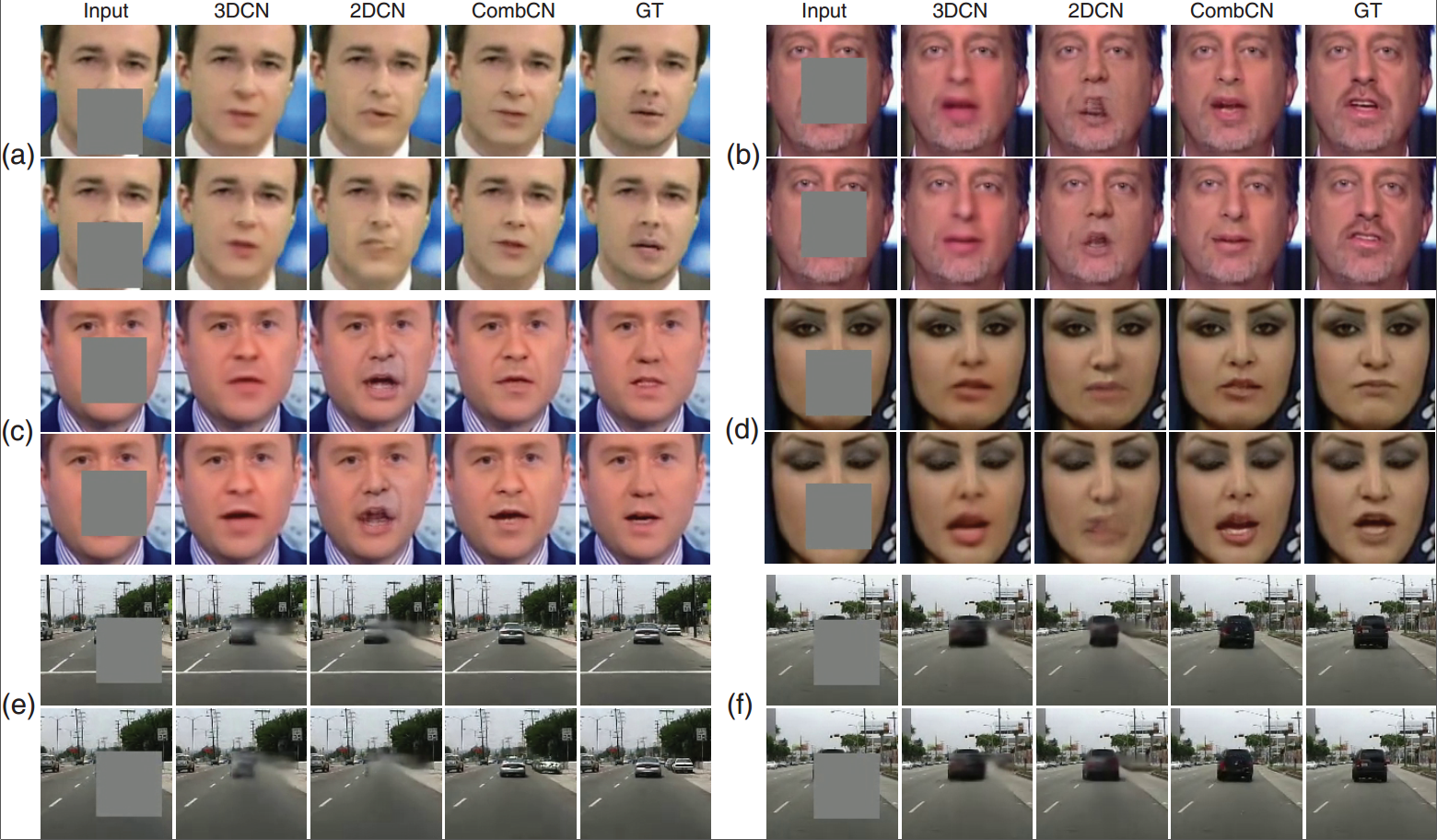}  
  \caption{Inpainted frames on datasets FaceForensics (a$\sim$d) and Caltech (e, f). In each panel, the two rows represent two frames of a video, and the five columns from left to right are input, results by 3DCN, 2DCN and CombCN, as well as the target ground truth. Better visual experience can be obtained in our accompanying supplemental materials.}\label{fig:rst-ff4k} 
\end{figure*}

To tackle the issue that 2DCN treats each frame independently without considering temporal coherence, we also inject the information from the output of 3DCN to CombCN. This is achieved by using two convolutional layers to extract two feature maps of the 3DCN output separately. The two feature maps are then added to the first and last layer of the same size in CombCN, serving as a temporal guidance. In this paper, since 3DCN works on videos of size $\frac{H}{2}\times\frac{W}{2}$, the combination takes place in the 2nd and the 15th layer. We compared the effectiveness of this combination setting with the basic 2DCN on successive frames. The experimental results illustrate temporal coherence can be well preserved when inpainting frames separately, as shown in Figs.~\ref{fig:rst-ff4k} and~\ref{fig:teaser}.
\paragraph{Training.} Let $G_i(V_{in}^k, M^k, I^{d,k}_{out}) = I^k_{out}$ denotes the CombCN in a functional form, where $V_{in}^k$, $M^k$ and $I^{d,k}_{out}$ are the $k$-th frame of the incomplete video $V_{in}$, mask video $M$ and the inpainted video $V^d_{out}$ by 3DCN. During training, we view a video data sample as a batch of images so that the data format can be well supported by the existing deep learning frameworks like TensorFlow. For a video data sample, the optimization goal is to minimize the mean of the $l_1$ norm of the difference between $I^k_{out}$ and $V_c^k$ ($k=1,2,...,F$). Specifically, the loss of CombCN is defined as:

\begin{align}\label{eq:l1-combcn}
& L^{CombCN}(V_{in}, M, V^{d}_{out}, V_c) \\ \nonumber
& = \sum_{k=1,2,...,F}\frac{\lVert M^k \odot (G_i(V_{in}^k, M^k, I^{d,k}_{out}) - V_c^k)\rVert}{F \cdot \lVert M^k \rVert}
\end{align}

In practice, we first pre-train 3DCN to converge, and then train CombCN with the pre-trained 3DCN model finetuned. This training strategy can lead to fast convergence of CombCN compared with training both sub-networks together from scratch. We also enable finetuning 3DCN in order to acquire lower loss compared with the strategy without finetuning. In this case, we jointly optimize the weighted sum of the two sub-network losses, i.e.
\begin{equation}\label{eq:total-loss}
L^{total} = L^{3DCN} + \alpha L^{CombCN}
\end{equation}
where $\alpha$ is a balancing parameter which is set to $1.0$ in our paper.
A detailed comparison of different training strategies are presented in Section~\ref{sec:exp-ts}\textit{Performance of variants of training strategy}. The experiment result shows that our training strategy outperforms the other two, i.e. pre-training 3DCN disabled and finetuning 3DCN disabled when training CombCN.
\section{Experimental Results}\label{sec:exp}

\subsection{Dataset and implementation details}
To validate our 3D-2D combined completion network, we tested on three datasets, FaceForensics~\cite{roessler2018faceforensics}, 300VW~\cite{chrysos2015offline} and Caltech~\cite{Dollar2012PAMI}.

The first two datasets contain 1,004 and 300 video clips with human faces respectively, where the faces are near-frontal pose and neutral expression change across frames. To further stress test our method, we also run on Caltech~\cite{Dollar2012PAMI}  which  contains 10 hours of video of size $640\times480$ taken from a vehicle driving through regular traffic in an urban environment. Compared with the face videos with obvious semantic structures, Caltech dataset is more challenging to an inpainting task.

In data preparation stage, we group every 32 frames from the original video clips into a data sample. Each frame is in $128^2$ resolution, which is generated in the following manner. For the face videos, i.e. FaceForensics and 300VW, we first crop the face out with a squared bounding box; while for Caltech dataset, we directly crop the central $480^2$ region out. The cropped region is then resized into $128^2$ resolution. For each dataset, we separate the whole data samples into training and validation sets and control their proportion $5:1$.

During training we randomly generate a hole across all frames in the $[0.375l,0.5l]$ pixel range and fill it with the mean pixel value of the training dataset, where $l$ is the frame size ($128$ in this paper). The range follows the same ratio as in~\cite{IizukaSIGGRAPH2017}. The position of the hole for a video data sample is identical for all of its frames. Based on these inputs, we first pre-train 3DCN to convergence and then train CombCN with the pre-trained 3DCN model jointly finetuned. The CombCN is trained with $100k$ iterations by an Adam optimizer, whose regression weight and learning rate are set to $0.01$ and $0.001$, respectively. Each iteration costs approximate 0.8s and it takes nearly 30 hours to complete the entire training. The detailed configuration of 3DCN and CombCN is illustrated in Fig.~\ref{fig:net3d}. The implementation is based on TensorFlow and the network training is performed on a single NVIDIA GeForce GTX 1080 Ti. 
\begin{figure*}[t!]
  \centering
  \includegraphics[width=0.92\linewidth]{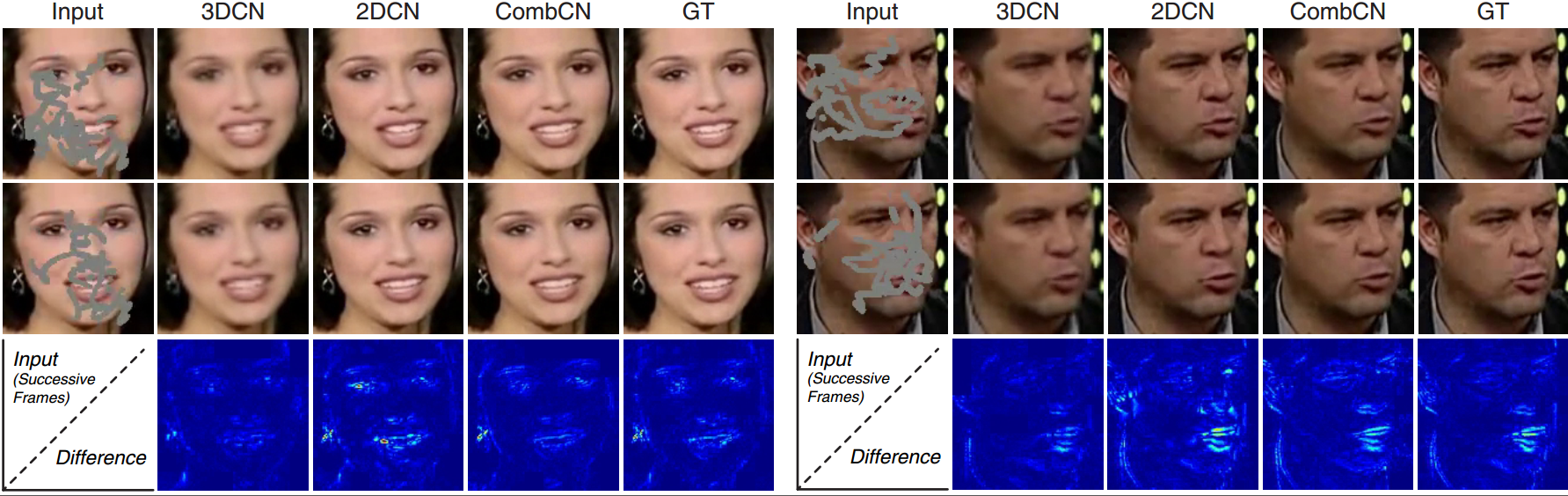}\\
  \caption{Inpainting results from videos with random holes. We visualize the differences between the two successive frames to illustrate the inter-frame consistency. It shows that our CombCN produced clearer results than 3DCN, and smoother results than 2DCN. Better visual experience can be obtained in our accompanying supplemental materials.}\label{fig:random-holes}
\end{figure*}

\subsection{Comparisons with existing methods}
Our results in comparison with those produced by 2DCN and 3DCN can be found in Figs.~\ref{fig:rst-ff4k} and~\ref{fig:random-holes}. More video results are also presented in our accompanying supplemental materials. 
\paragraph{Ours vs. 2DCN.}
We first compare our method with a state-of-the-art image inpainting method~\cite{IizukaSIGGRAPH2017}, which is applied to each video frame independently. Such an image inpainting method is expected to produce higher quality result, but when applying it to each video frame, we found that the quality of frames are unstable and annoying visual artifacts like flicker are present. This is because it lacks the mechanism to preserve the spatial-temporal coherence of the video. For example, in the 3rd column of Fig.~\ref{fig:rst-ff4k}(a)(d), the algorithm produces high quality results on the first frame but fails in the second frame, making the mouths missing. This circumstance also occurs in the Caltech dataset as shown in Fig.~\ref{fig:rst-ff4k}(e), where the car disappears unreasonably in the second frame.  In addition, if there is motion in video, 2DCN may also produce blurry or distorted results as illustrated in Fig.~\ref{fig:rst-ff4k}(b)(c). In contract, our CombCN produces reasonable and stable results (the 4th column) across frames. This temporal coherence ensures the pleasant user experience when the video being played. 
\paragraph{Ours vs. 3DCN.}
We also compare our results with the low resolution output achieved by 3DCN. They are listed in the 2nd column in Fig.~\ref{fig:rst-ff4k}. Due to the usage of 3D CNN in addition to working on low resolution, the inpainted frames contain significant blurry artifacts so that details cannot be well reconstructed. For example in Fig.~\ref{fig:rst-ff4k}(b)(c), the teeth are almost missing in the inpainted frames by 3DCN. However, unlike the results by 2DCN, they have smooth transition across frames. In comparison, our approach preserves temporal coherence and details simultaneously. The final $l_1$ losses\footnote{The value of $l_1$ loss in this paper has been normalized so that it is equal to the mean error for each pixel. Its value range is $[0, 255]$.} by 3DCN, 2DCN and CombCN are listed in Table.~\ref{tab:convergence-error} (top). 
\begin{table}[!t]
\centering
\resizebox{2.5in}{!}{
\begin{tabular}{c|c|c|c}
\hline
             & 3DCN  & 2DCN  & CombCN (ours) \\ \hline
FaceForensics & 7.18  & 6.77  & \textbf{6.27}          \\
Caltech       & 11.91 & 11.16 & \textbf{9.56}          \\ \hline
\end{tabular}
}\\ 
\resizebox{2.7in}{!}{
\begin{tabular}{c|cc|cc|c}
\hline
       & V-1   &   V-2   & T-1   & T-2 & our method \\ \hline
3DCN   &  9.51  &  11.56 & 6.30 & 9.28 & \textbf{4.45}  \\
CombCN &  6.39  &   8.13 & 5.18 & 6.31  & \textbf{4.20}  \\ \hline
\end{tabular}}
\caption{Final $l_1$ losses. Top: the losses of 3DCN, 2DCN and CombCN of datasets FaceForensics and Caltech. Bottom: the losses of 3DCN and CombCN in 300VW dataset, based on variants of 3DCN (V-1, V-2) and training strategy (T-1, T-2), in comparison with our method.}
\label{tab:convergence-error}
\end{table}

\paragraph{Random holes.} Our system can also be easily applied to the inpainting task with random holes in validation/testing phase, even in the case that holes are distinct for the input frames. Note that in this case, though working in a lower resolution, 3DCN can further reveal its power to fill the holes in a temporally consistent manner, because it can take advantage of the pixels of the non-hole regions in the contiguous frames. However, distinct holes make it more challenging for 2DCN to keep the inter-frame consistency. As a result, our CombCN combines the two benefits from the two sub-networks and is able to produce clearer and smoother results. Fig.~\ref{fig:random-holes} shows two examples of video inpainting with random holes.

\subsection{Ablation studies}

To discover the vital elements in the success of our proposed model for video inpainting, we made two groups of variants of our method. They are based on modifications of the 3DCN structure and the training strategy separately. The final losses of all variants and our method are listed in Table.~\ref{tab:convergence-error} (bottom). These ablation studies were conducted on 300VW dataset. 

\subsubsection{Performance of variants of 3DCN}
To investigate the influence of 3DCN to the final results by CombCN, we first modified the structure of 3DCN to its two variants V-1, V-2 as below.
\begin{description}
  \item[V-1.] \textbf{Feed 3DCN with videos in lower resolution.}\newline
  In this experiment, we fed the 3DCN with a lower resolution version of the original video, i.e. setting down-sample rate $r=4$ to produce a $32^3$ video $\tilde{V^d_{in}}$. Accordingly, we changed the combination layers in CombCN to the first and last feature maps of size $32^2$ instead of $64^2$. In this setting, the 3DCN produces more blurry frames while the temporal coherence is rarely lost. Our experiment shows that the convergence errors of 3DCN and CombCN are 9.51 and 6.39, while our baseline model produces the corresponding errors 4.45 and 4.20.
  \item[V-2.] \textbf{Involve down-sampling in time axis in 3DCN.} \newline
  We alternatively modified our basic 3DCN to allow strided convolutions across time-axis. So the frame number of the feature maps in the 2nd and 4th convolutional layers are down-sampled, becoming $\frac{F}{2}$ and $\frac{F}{4}$ respectively. The deconvolutional layers are also modified to support up-sampling in time-axis. In this setting, the temporal coherence of the inpainted frames by 3DCN are less preserved while the extracted features become more compact. Our experiment shows that visually this setting does not obviously down-grade the performance of CombCN, and the convergence errors increase to 11.56 for 3DCN and 8.13 for CombCN.
\end{description}



\subsubsection{Performance of variants of training strategy} \label{sec:exp-ts}
As afore-mentioned, our experimental results were produced by a CombCN trained with a fine-tuned pre-trained model of 3DCN. To investigate the performance of this training strategy, we further compared two other strategies T-1 and T-2 as follows.
\begin{figure}
  \centering
  \includegraphics[width=0.9\linewidth]{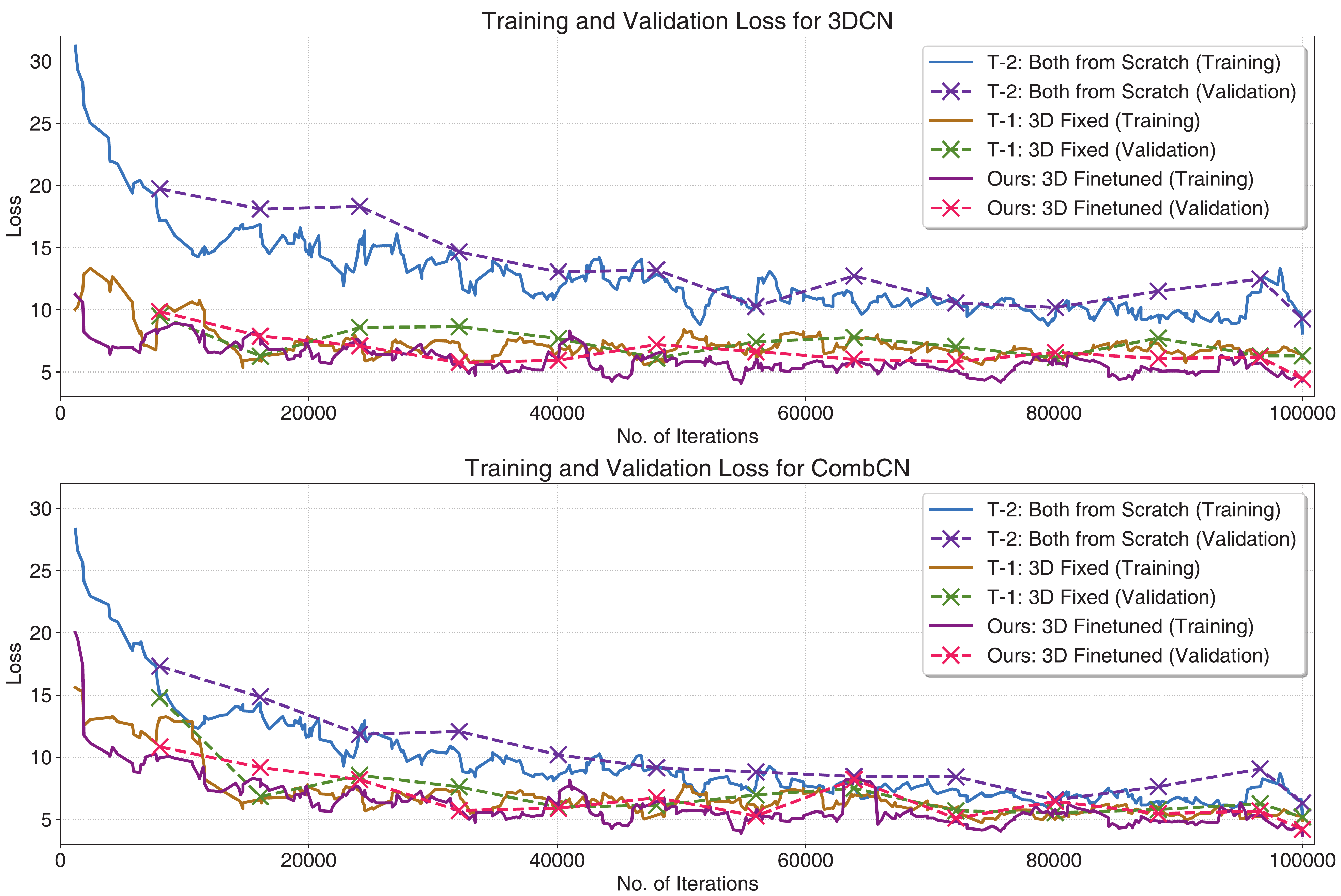} 
  \caption{Training and validation loss of variants of training strategy for 3DCN (top) and CombCN (bottom).}\label{fig:training-strategy-error} 
\end{figure}
\begin{description}
  \item[T-1.] \textbf{Pre-train 3DCN, then train CombCN without fine-tuning it.} \newline
  We first disable fine-tuning the pre-trained model of the 3DCN when training the CombCN. In this setting, the pre-trained 3DCN produces inpainted video frames in low resolution and they are directly fed forward to the CombCN. Due to lack of parameter updating, the loss of 3DCN keeps fluctuating without dropping, while the loss of the CombCN drops to convergence after nearly $20k$ iterations. After nearly $100k$ iterations, the loss approaches the value that is achieved by the strategy of fine-tuning enabled. We plot the two losses of 3DCN and CombCN in $100k$ iterations labelled by "T-1: 3D Fixed" in Fig.~\ref{fig:training-strategy-error}.
  \item[T-2.] \textbf{Train 3DCN and CombCN jointly from scratch.} \newline
  We further remove the stage of pre-training 3DCN, and train 3DCN and CombCN jointly from scratch. Compared with our method and T-1, due to the lack of rich informative guidance produced by 3DCN, it takes considerable iterations (over $50k$) for CombCN and 3DCN to converge. Furthermore, the convergence losses of the two sub-networks are also higher, i.e. 9.28 for 3DCN and 6.31 for CombCN. The result reveals the capability of our current training strategy in this paper.
\end{description}


\subsection{Limitations}

\begin{figure}[!t]
  \centering
  \includegraphics[width=0.82\linewidth]{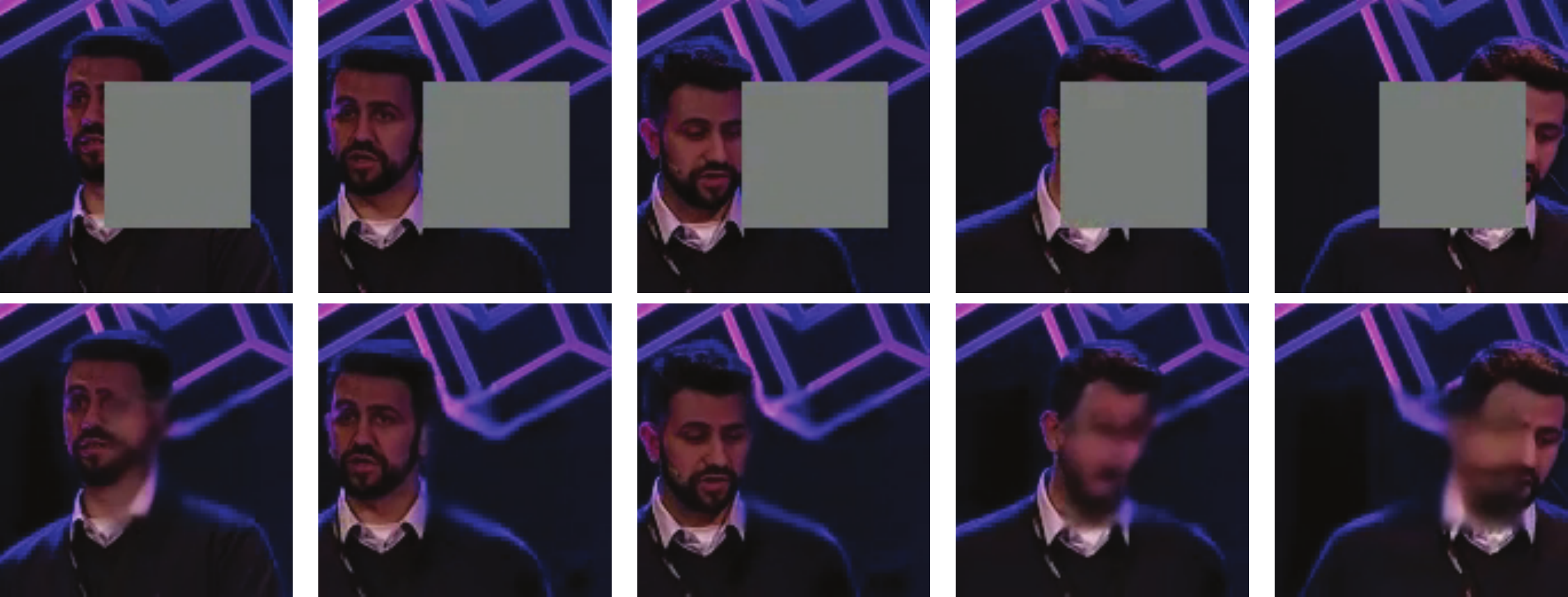}\\ 
  \caption{Failure case. Top: five frames with holes as input. Bottom: inpainted results by CombCN.}\label{fig:limitation}
\end{figure}

Our approach can handle videos with unstructured information and motion in most cases. However, it may still fail if the test video has a severe variation from the training data. Furthermore, inferring temporal coherence relies on 3DCNN so that large motion cannot be easily captured due to the limitation of the size of receptive field. Fig.~\ref{fig:limitation} illustrates an example of our failure case, where the test video displays a man with large motion and was captured in a different view setting (far from the face). As a result, our approach produces an unreasonable face in the 4th frame. We believe using an optical flow and LSTM based solution as in \cite{Lai-ECCV-2018,ren2016look} could be a potential idea for this problem. 

Moreover, unlike state-of-the-art image inpainting approaches commonly involving GAN to synthesize more vivid results, we only use $l_1$ loss in our paper, which may potentially limit the power of our combination idea. We will leave it as a future work about how to integrate GANs to our 3DCN and CombCN. 
\section{Conclusion}\label{sec:conclusion} 
We have presented an end-to-end framework for video inpainting through a joint 2D-3D CNN which contains a temporal structure inference network and spatial detail recovering network. Our method can fill regular or random holes across frames to produce plausible results. These results show that our method significantly improves the performance of existing methods. We also believe this architecture has potentials to be applied to other video generation tasks. 

\section{Acknowledgements}
The work was partially supported by Shenzhen Fundamental Research Fund under Grant No. KQTD2015033114415450, and "The Pearl River Talent Recruitment Program Innovative and Entrepreneurial Teams in 2017" under grant No. 2017ZT07X152. We also thank the reviewers, Ms. Chang Li from University of Washington and Mr. Zhangyang Xiong from CUHK (Shenzhen) for their constructive comments and criticism of the manuscript. 

\bibliography{egbib}
\bibliographystyle{aaai}

\end{document}